\documentclass{article}

\usepackage[final]{neurips_2024}

\usepackage[utf8]{inputenc} 
\usepackage[T1]{fontenc}    
\usepackage{hyperref}       
\usepackage{url}            
\usepackage{booktabs}       
\usepackage{amsfonts}       
\usepackage{nicefrac}       
\usepackage{microtype}      
\usepackage{xcolor}         

\usepackage{caption}
\usepackage{subcaption}

\usepackage{tabu}                      
\usepackage{booktabs}                  
\usepackage{lipsum}                    
\usepackage{mwe}                       

\usepackage{mathptmx}
\usepackage{tikz}
\usepackage{graphics}
\usepackage{color}
\usepackage{float}
\usepackage{epsf}
\usepackage{caption}
\usepackage{amsmath,amsfonts}
\usepackage{algorithm}
\usepackage{algpseudocode}

\usepackage{bm}
\usepackage{multirow}
\usepackage{booktabs}                  
\usepackage{lipsum}                    
\usepackage{mwe}                       

\title{Evaluating Loss Landscapes \\ from a Topology Perspective}

\author{
  Tiankai Xie\thanks{Equal contribution.} \\
  Arizona State University \\
  \texttt{txie21@asu.edu} \\
  \And
  Caleb Geniesse$^*$ \\
  Lawrence Berkeley National Lab \\
  \texttt{cgeniesse@lbl.gov} \\
  \And
  Jiaqing Chen$^*$ \\
  Arizona State University \\
  \texttt{jchen501@asu.edu} \\
  \And
  Yaoqing Yang \\
  Dartmouth College \\
  \texttt{yaoqing.yang@dartmouth.edu} \\
  \And
  Dmitriy Morozov \\
  Lawrence Berkeley National Lab \\
  \texttt{dmorozov@lbl.gov } \\
  \And
  Michael W.~Mahoney \\
  ICSI, LBNL, and UC Berkeley  \\ 
  \texttt{mmahoney@stat.berkeley.edu} \\
  \And
  Ross Maciejewski \\
  Arizona State University \\
  \texttt{rmacieje@asu.edu} \\
  \And 
  Gunther H.~Weber \\
  Lawrence Berkeley National Lab \\
  \texttt{ghweber@lbl.gov} \\
}

\begin{document}

\maketitle

\begin{abstract}
  Characterizing the loss of a neural network with respect to model parameters, i.e., the \emph{loss landscape}, can provide valuable insights into properties of that model. Various methods for visualizing loss landscapes have been proposed, but less emphasis has been placed on quantifying and extracting actionable and reproducible insights from these complex representations. Inspired by powerful tools from topological data analysis (TDA) for summarizing the structure of high-dimensional data, here we characterize the underlying \emph{shape} (or topology) of loss landscapes, quantifying the topology to reveal new insights about neural networks. To relate our findings to the machine learning (ML) literature, we compute simple performance metrics (e.g., accuracy, error), and we characterize the local structure of loss landscapes using Hessian-based metrics (e.g., largest eigenvalue, trace, eigenvalue spectral density). Following this approach, we study established models from image pattern recognition (e.g., ResNets) and scientific ML (e.g., physics-informed neural networks), and we show how quantifying the shape of loss landscapes can provide new insights into model performance and learning dynamics. 
\end{abstract}

\section{Introduction}

Given the important role that the loss function plays during learning, examining it with respect to a neural network's weights---by visualizing the so-called \emph{loss landscape}---can provide valuable insights into both network architecture and machine learning (ML) dynamics~\citep{martin2021implicit,martin2021predicting,yang2022evaluating,yang2021taxonomizing,zhou2023three}. Indeed, the loss landscape has been essential for understanding certain aspects of deep learning, including, but not limited to, test accuracy, robustness of transfer learning~\citep{djolonga2021robustness}, robustness to out-of-distribution detection~\citep{yang2022generalized}, robustness to adversarial attack~\citep{kurakin2016adversarial}, and generalizability~\citep{cha2021swad}. There are two popular approaches to generating the loss landscape for a given neural network model. Initial efforts to visualize the loss landscape relied on sampling random orthogonal vectors and projecting weights onto the plane spanned by these random vectors~\citep{goodfellow2014qualitatively, li2018visualizing}. More recently, ~\citet{yao2020pyhessian} proposed using directions based on the Hessian, wherein the first two most important Hessian eigenvectors are used to capture more meaningful changes in the loss function. In both approaches, a neural network's parameters are perturbed along each direction, and the loss is re-evaluated at each of these positions. 

While both approaches have provided valuable insights, \emph{loss landscape visualization} (no matter which method was used) is often limited to just that---visualization. In other words, loss landscapes, once created, are often simply visually explored or qualitatively compared. It is less clear how to meaningfully measure or quantitatively relate these landscapes to features of the model's underlying architecture or to properties inherent to the learning process. Indeed, examining and quantifying a loss landscape---which is inherently high-dimensional, with as many dimensions as the number of parameters in the model---is challenging to do, especially when using two-dimensional views and qualitative observations alone.

To provide a more quantitative approach to understanding and using loss landscapes, here we show how topological data analysis (TDA) can be used to quantify and extract (quantitative) insights based on the topology (or shape) of those landscapes. We first compute loss landscapes using either random projections or Hessian-based directions and explore four different representations, including one image data representation (where the loss is stored as pixels) and three unstructured grid representations (where the loss is stored on the vertices of a graph). We then apply two methods from TDA, namely, the merge tree~\citep{contour_tree, heine2016survey} and persistence diagram~\citep{edelsbrunner2008persistent}, to quantify and compare different loss landscapes. We quantify these structures by measuring the number of saddle points and average persistence, respectively, and we compare our results with state-of-the-art methods for evaluating model performance, as well as with more recent methods for evaluating the local geometry of loss landscapes based on the Hessian.

\section{Background on TDA}
\label{sec:background}

Topological data analysis (TDA) aims to reveal the global underlying structure of data. It is particularly useful for studying high-dimensional data or functions, where direct visualization is impossible. Here, we leverage ideas and algorithms from TDA to study the structure of the loss function, i.e., the so-called loss landscape. In the context of a loss function, we are interested in the number of minima (i.e., unique sets of parameters for which the loss is locally minimized) and how ``prominent'' they are (i.e., measuring how many other sets of neighboring parameters have a higher loss than the parameter sets that locally minimize the loss function). Such information can be obtained from the merge tree and persistence diagram (i.e., captured by the $0$-dimensional persistent homology). 

A \emph{merge tree} \citep{contour_tree, heine2016survey} tracks connected components of sub-level sets $L^-(v) = \{ x \in D; x \le v\}$ as a threshold, $v$, is increased. Note, the merge tree can track either sub-level or super-level sets, but here we are interested in characterizing loss functions and their minima, so we focus on sub-level sets. In this case, as $v$ increases, new connected components form at local minima and later merge with neighboring connected components (other local minima) at saddles. The merge tree encodes these changes in the loss landscape as nodes in a tree-like structure, where local minima are represented by degree-one nodes (connected to other local minima through a single saddle point), and the saddle points connecting different minima are represented by degree-three nodes (each connecting two local minima and one other saddle point). 

A \emph{persistence diagram} represents features (i.e., branches in the merge tree) as points in a two-dimensional plane. The horizontal axis corresponds to the birth of each feature—which is the value of the minimum at which it first appears. The vertical axis corresponds to the death of each feature—which is the value of the saddle where it merges into a more persistent feature. The distance between a point and the diagonal line $y = x$ encodes the persistence of the feature—which is a measure of how long the feature lasts in the filtration (i.e., as the threshold, $v$, is increased). Such a metric captures information about the ruggedness of the landscape, and for example, the depth of local valleys and the height of the barriers between them. Here, we quantify the average persistence by computing the distance between each persistence pair and the diagonal and then taking the average.

\section{Empirical Results}
\label{sec:experiments_results}

In this section, we provide a summary of our main empirical results. We study established models from image pattern recognition (e.g., ResNets) and scientific ML (e.g., physics-informed neural networks). We first show how removing residual connections from ResNet changes the shape of the loss landscape. We then show how the loss landscape changes for a physics-informed neural network as a physical parameter is varied and optimization begins to fail. In both experiments, we quantify the merge tree and persistence diagrams, and we relate our results to model performance and loss landscape metrics like the top Hessian eigenvalue and the Hessian trace.

\subsection{Image Pattern Recognition}
\label{sec:experiments_results_resnet}

In our first experiment, we explore image pattern recognition using ResNet-20 trained on CIFAR-10~\citep{yao2020pyhessian}. Specifically, we look at (and quantify) loss landscapes before and after adding residual connections. Recent work shows that the residual connections are related to the ``smoothness'' of the loss landscape~\citep{li2018visualizing,yao2020pyhessian}. Here, we aim to verify this and further characterize how the residual connections in ResNet-20 change the underlying loss landscape. Note, we removed the residual connections from ResNet-20 before training. The accuracy of ResNet-20 without residual connections (90\% average accuracy across four random seeds) was slightly lower than ResNet-20 with residual connections (92\% average accuracy across four random seeds).

\begin{figure*}[tbh]
    \centering	
    \includegraphics[width=1.00\columnwidth]{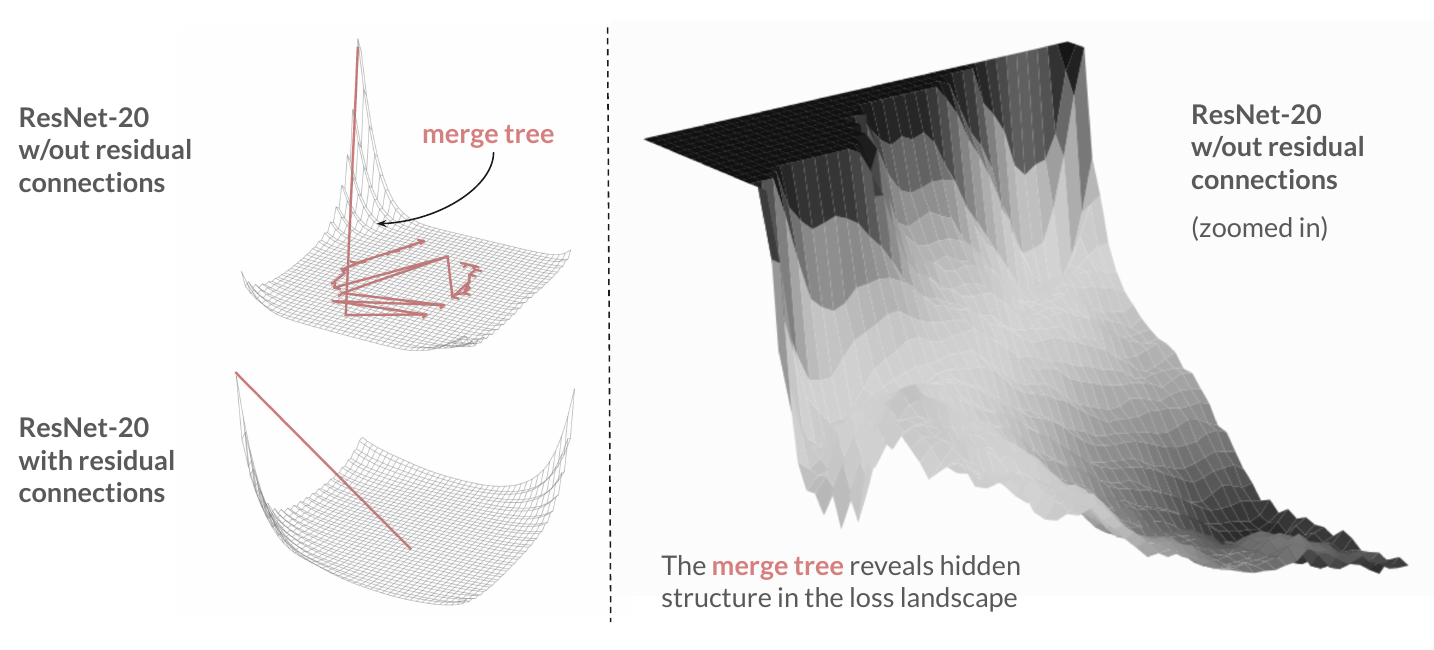}
    \caption{Visualizing loss landscapes for ResNet-20. Compare with~\citet{yao2020pyhessian}.}
    \label{fig:resnet_random}
\end{figure*}

In Fig.~\ref{fig:resnet_random}, we show loss landscapes for ResNet without and with residual connections. We observe a much more complicated structure in the loss landscape for ResNet-20 without residual connections, revealed by the more complicated branched structure of the merge tree. Interestingly, this complicated structure is not immediately visible in the loss landscape itself. On the right, we show the same landscape for ResNet-20 without residual connections, after clipping outlier values to ``zoom'' into the smaller scale structure. The misleading visualization on the left highlights a unique benefit of our approach. That is, the merge tree can capture interesting structure across different scales by default, requiring less manual fine-tuning of visualization parameters. Comparatively, we observe a much smoother loss landscape for ResNet-20 with the residual connections, confirmed by the simpler structure of the merge tree (i.e., a single minima and no saddle points). Interestingly, these observations agree with previous findings that adding residual connections to ResNet results in a ``smoother'' loss landscape, thereby improving generalization~\citep{li2018visualizing}. 

In Fig.~\ref{fig:experiments_resnet20_scatter_plots} (in the appendix), we further verify these observations numerically by showing how our TDA-based metrics relate to ML-based metrics. These plots provide additional insights beyond the qualitative differences in the loss landscapes we observed, revealing that the number of saddle points in the merge tree increases and the average persistence decreases when the residual connections are removed from ResNet-20. Looking across the different columns, we observe an inverse relationship between the number of saddle points in the merge tree and the ML-based metrics, but a direct relationship between the average persistence and the same ML-based metrics. Together, these results provide additional insight into how changing the architecture of a neural network like ResNet-20 can result in a “smoother” (and thereby easier to optimize) loss landscape.

\subsection{Physics-Informed Neural Networks}
\label{sec:experiments_results_pinn}

\begin{figure*}[!b]
    \centering	
    \includegraphics[width=1.00\columnwidth]{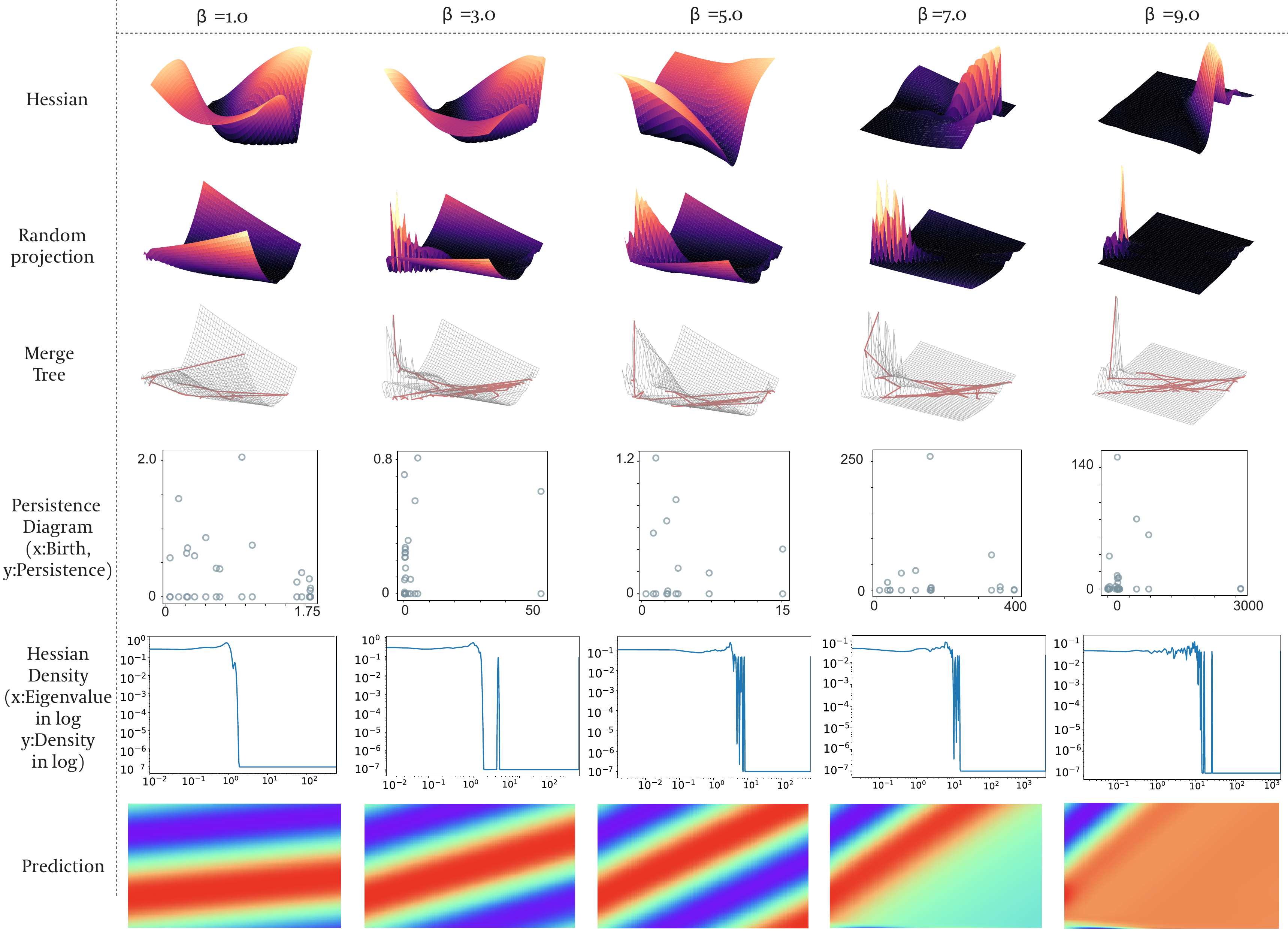}
    \vspace{-0.8em}
    \caption{Visualizing the failure modes of PINNs. Compare with~\citet{krishnapriyan2021characterizing}. 
    }
    \label{fig:pinn}
\end{figure*}

In our second experiment, we look at a set of physics-informed neural network (PINN) models trained to solve increasingly difficult convection problems. Specifically, we consider the one-dimensional convection problem, a hyperbolic partial differential equation that is commonly used to model transport phenomena:
\begin{align}\label{eq:PINN_1}
    \frac{\partial u}{\partial t} + \beta \frac{\partial u}{\partial x} = 0,\ \ x \in \Omega,\ \ t \in [0, T] 
\end{align}
\vspace{-4mm}
\begin{align}\label{eq:PINN_2}
    u(x, 0) = h(x),\ \ x \in \Omega ,
\end{align}
where $\beta$ is the convection coefficient and $h(x)$ is the initial condition. The general loss function for this problem is
\vspace{-3mm}
\begin{align}\label{eq:PINN_3}
    L(\theta) = \frac{1}{N_{u}}\sum_{i=1}^{N_{u}} (\hat{u}-u_0^i)^2 + \frac{1}{N_{f}}\sum_{i=1}^{N_{f}} \lambda_i (\frac{\partial \hat{u}}{\partial t} + \beta \frac{\partial \hat{u}}{\partial x})^2 + L_B ,
\end{align}
where $\hat{u} = N N (\theta, x, t)$ is the output of the NN, and $L_B$ is the boundary loss. The goal of this case study is to investigate the PINN’s soft regularization and how it helps (or fails to help) the optimizer find optimal solutions to seemingly simple convection problems. As shown in~\citet{krishnapriyan2021characterizing}, increasing the wave speed parameter, $\beta$, can make it harder for the PINN to find a reasonable solution. This difficulty has been linked to changes in the loss landscape, which becomes increasingly complicated, such that optimizing the model becomes increasingly difficult.

In Fig.~\ref{fig:pinn}, we show that increasing the wave speed indeed results in a more complicated loss landscape. In the bottom row, we show the spatiotemporal patterns predicted by the PINN models. Note, increasing $\beta$ makes the problem harder to solve, and here we can see that our PINN models fail to accurately predict the pattern around $\beta = 7.0$. Looking at the first three rows, we observe increasingly complex loss landscapes as $\beta$ increases, in both versions of the loss landscape and further revealed by the more complicated and branched structure of the merge tree (i.e., with many more minima and saddle points). The change in complexity of the landscape is further supported by higher persistence values in the persistence diagrams. We also relate our observations to the Hessian Density, which suggests that the volume and complexity of the loss landscape increases with $\beta$. Together, these different views give us a more holistic understanding of how the loss landscape changes as $\beta$ increases. Interestingly, these observations agree with previous findings that increasing $\beta$ results in a more complicated loss landscape, corresponding to a harder optimization problem~\citep{krishnapriyan2021characterizing}. We further verify these observations numerically in Fig.~\ref{fig:experiments_pinn_scatter_plots} (in the appendix). These plots provide additional insights beyond the qualitative changes in the loss landscapes we observed, confirming that the number of saddle points in the merge tree (left column) increases, along with the average persistence (right column), as the value of $\beta$ increases. We observe similar trends in the Absolute Error, Top-1 Hessian Eigenvalue, and Hessian Trace as $\beta$ increases. Together, these results reaffirm previous findings and provide new insights into the failure modes of PINNs, revealing that the topology of the loss landscape becomes significantly complex as PINNs start to fail.

\section{Conclusion and Future Work}
\label{sec:conclusion_future_work}

In this work, we study neural network loss landscapes through the lens of TDA. Specifically, we capture the underlying shape of loss landscapes using merge trees and persistence diagrams. By quantifying these topological constructs, we reveal new insights about the landscapes. Furthermore, we explore the relationship between our TDA-based metrics and relevant traditional ML metrics. For ResNet models, we find that the number of saddle points is inversely related to the average persistence and to the other ML-based metrics. For PINN models, we find that the number of saddle points and average persistence increase together along with the other ML-based metrics. These relationships reflect the curvature and sharpness of the landscape, which in turn strongly impacts the model's performance and generalization abilities. Note, in this work we only show the $0$-dimensional persistence diagram, which is exactly what the branches in the merge tree encode. Since here our original focus was on extracting merge trees, we decided to limit our analysis to these lower-dimensional features. We leave the analysis of higher-dimensional holes for future work.

Moreover, since the merge tree and persistence diagram can be computed for arbitrary-dimensional spaces, our generalized and scalable approach opens up the door to studying loss landscapes in higher dimensions. In the future, we hope to extend our approach to characterize higher-dimensional loss landscapes, under the hypothesis that more information (hidden in these additional dimensions) could perhaps provide new insights into the loss function and properties relating neural network architecture to learning dynamics. One simple way to do this would be to sample along more directions. So, for example, we could construct a subspace based on the top 3 to 10 Hessian eigenvectors. While this would still be far from the potentially billions of dimensions in the true high-dimensional loss landscapes of modern ML models (with potentially billions of parameters), we expect that there exists a much lower-dimensional manifold upon which the interesting variation can be observed.

\section{Acknowledgments}

This work was supported by the U.S. Department of Energy, Office of Science, Advanced Scientific Computing Research (ASCR) program under Contract Number DE-AC02-05CH11231 to Lawrence Berkeley National Laboratory and Award Number DE-SC0023328 to Arizona State University (“Visualizing High-dimensional Functions in Scientific Machine Learning”). This research used resources at the National Energy Research Scientific Computing Center (NERSC), a U.S. Department of Energy, Office of Science, User Facility under NERSC Award Number ASCR-ERCAP0026937.

\bibliographystyle{plainnat}
\bibliography{neurips_2024}


\newpage 
\appendix

\label{sec:appendix}

\section{Appendix}
\label{apd:first}

\subsection{Loss Landscapes}
\label{apd:methods_loss_landscapes}

To compute a loss landscape, we need to project the high-dimensional loss function into a two-dimensional space. To construct this space, we therefore need to sample two orthonormal vectors. We consider two different ways to sample these vectors (e.g., random directions, Hessian-based directions), but the rest of the procedure for sampling the loss function in the resulting spaces is the same regardless of how the directions were chosen. 

The first (naive) way to define the lower-dimensional subspace is to sample two random vectors. In high dimensions, they will most likely be nearly orthogonal. Here, we use PyTorch to randomly sample two vectors having the same size as the number of parameters in the model.

The second way to define the lower-dimensional subspace is to use the top two Hessian eigenvectors. Since the eigenvectors associated with the largest eigenvalues correspond to the directions of most variation in the loss function, sampling along these directions can often reveal more interesting surfaces. Here, we use PyHessian \citep{yao2020pyhessian}, a PyTorch library for efficiently calculating Hessian information using (randomized) numerical linear algebra. 
 
Given the two orthonormal directions, we follow the approach outlined by \citet{li2018visualizing}. More formally, we sample the trained model at discrete positions in the subspace defined by the two directions and evaluate the loss $\mathcal{L}$ as follows:
\begin{equation}
  \label{eq:projection_general}
  f(\alpha_1, \alpha_2) = \mathcal{L} (\theta + \alpha_1\delta_1 + \alpha_2\delta_2)  ,
\end{equation}
where $(\alpha_1,\alpha_2)$ are the coordinates in the two-dimensional subspace, $\delta_1$ and $\delta_2$ correspond to the first and second direction defining that subspace, and $\theta$ is the original model. As such, each coordinate corresponds to a perturbed model, and we store the loss for that model, such that the collection of loss values forms the two-dimensional loss landscape. 

Given a two-dimensional loss landscape, we can represent the sampled points as \emph{image data}, where each pixel represents the loss of a perturbed model, or as an \emph{unstructured grid}, where each vertex in the grid is associated with a scalar loss value. For the unstructured grid, we need to define the spatial proximity (or connectivity) of vertices in the grid based on the similarity of their coordinates. This will allow us to characterize how the loss changes throughout the landscape (i.e., as parameters are perturbed from one vertex to the next).

Here we explored both triangulations (e.g., Delaunay, Gabriel graph) and neighbor-based graph approaches. We found that a scalable, approximate nearest neighbor algorithm \citep{dong2011efficient} yielded good results while also being much faster. Specifically, we use a $k$-nearest neighbor graph, where each point is connected to the $k$ most similar points, and we use $k = 8$, such that the connectivity is similar to the spatial proximity of pixels in an image (i.e., left, right, top, bottom, and all four~corners).

\subsection{Topological Data Analysis}
\label{sec:methods_tda}

We perform topological data analysis (TDA) to extract and summarize the most important features of the loss landscapes. We then further quantify the results from TDA to provide new metrics that can be related to things like model performance. Specifically, we compute the merge tree and the $0$-dimensional persistence diagram. Recall that the branches in the merge tree are equivalent to the $0$-dimensional features~\citep{edelsbrunner2008persistent}. However, compared to the persistence diagram which describes the life span of each component, the merge tree provides additional information including which component each component merges into. The merge tree also provides a nice intuitive visualization of the changes in the loss landscape. To quantify the merge tree, we count the number of saddle points and the number of minima. To quantify the persistence diagram, we measure the average persistence by computing the distance between each persistence pair and the diagonal and then taking the average.

Note, we compute the merge tree and persistence diagram for each loss landscape using the Topology ToolKit (TTK)~\citep{ttk2021overview}. Further analysis and visualization of the resulting merge trees and persistence diagrams was performed in Python.

\newpage
\section{Supplemental Figures}
\label{apd:second}

\vspace{24mm}
\begin{figure}[h]
    \centering	
    \includegraphics[width=0.70\columnwidth]{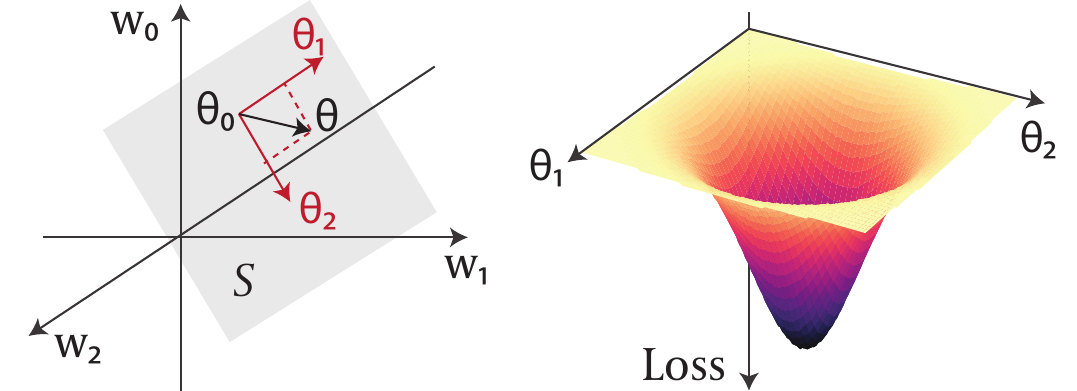}
    \caption{Illustration of generating the two-dimensional loss landscape. 
    In this work, to compute a loss landscape, we first construct a subspace defined by two vectors, $\theta_1$ and $\theta_2$. Within this subspace, we perform model interpolation and evaluate the corresponding loss values.
    }
    \label{fig:2D-projection}
\end{figure}

\vspace{20mm}
\begin{figure*}[th]
    \centering	
    \includegraphics[width=1.00\columnwidth]{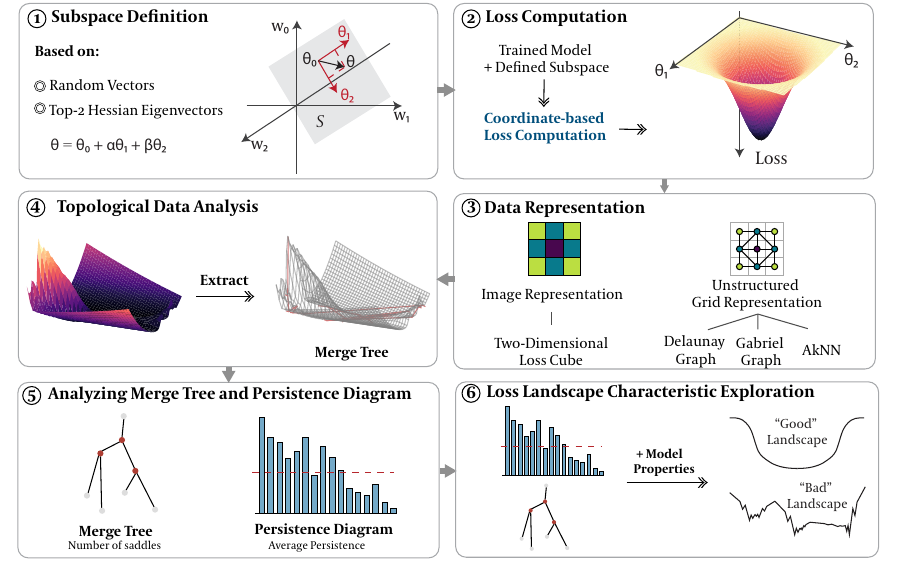}
    \caption{Our loss landscape analysis pipeline.
    The pipeline includes six stages: 
    (1) \emph{Subspace Definition}: Define the loss landscape subspace using random-based or Hessian-based directions.
    (2) \emph{Loss Computation}: Calculate loss values for coordinate locations using our coordinate-based loss computation method.
    (3) \emph{Data Representation}: Transform the loss landscape into suitable data structures for TDA.
    (4) \emph{Topological Analysis}: Extract the merge tree and persistence diagram.
    (5)~\emph{Quantitative Evaluation}: Quantify the merge tree and persistence diagram.
    (6) \emph{Loss Landscape Property Evaluation}: Relate the TDA-based metrics to loss landscape properties. 
    }
    \label{fig:analytics_framework}
\end{figure*}

\newpage
\begin{figure*}[th]
  \centering
  \includegraphics[width=\textwidth]{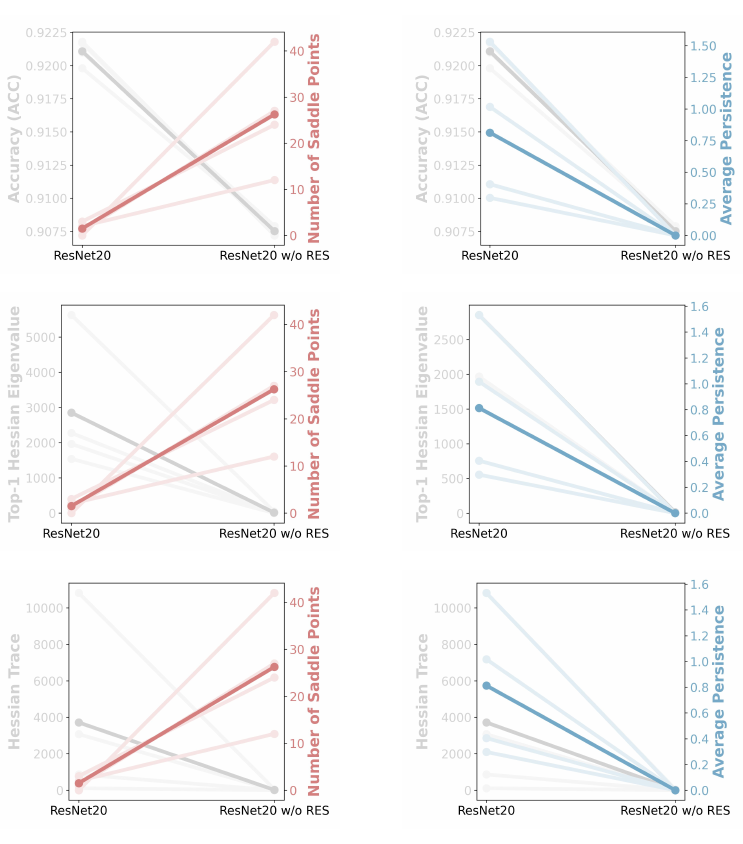}
  \caption{Quantifying the loss landscape for ResNet-20, with and without residual connections. We trained each version of the model four separate times, each time using a unique random seed (e.g., \texttt{0}, \texttt{123}, \texttt{123456}, and \texttt{2023}). Here, we numerically verify the observations we made in Fig.~\ref{fig:resnet_random} and provide additional insights based on persistence diagrams (not shown). We quantified the merge tree and persistence diagram by counting the number of saddle points and computing the average persistence, respectively. We compare our results with traditional machine learning metrics, including the Accuracy, Top-1 Hessian Eigenvalue, and Hessian Trace. Here, we show the relationship between those ML-based metrics and our two TDA-based metrics. These plots provide additional insights beyond the qualitative differences in the loss landscapes we observed in Fig.~\ref{fig:resnet_random}, confirming that the landscapes for ResNet-20 models without residual connections correspond to merge trees with a higher number of saddle points (left column). In contrast, we see that these models (without residual connections) display a lower average persistence (right column). We observe an inverse relationship between the number of saddle points in the merge tree and the ML-based metrics, but a direct relationship between the average persistence and the same ML-based metrics. Together, these results provide insight into how changing the architecture of a neural network like ResNet-20 (i.e., by adding residual connections) can result in a ``smoother'' (and thereby easier to optimize) loss landscape.
  }
  \label{fig:experiments_resnet20_scatter_plots}
\end{figure*}

\newpage
\begin{figure*}[th]
  \centering
  \includegraphics[width=\textwidth]{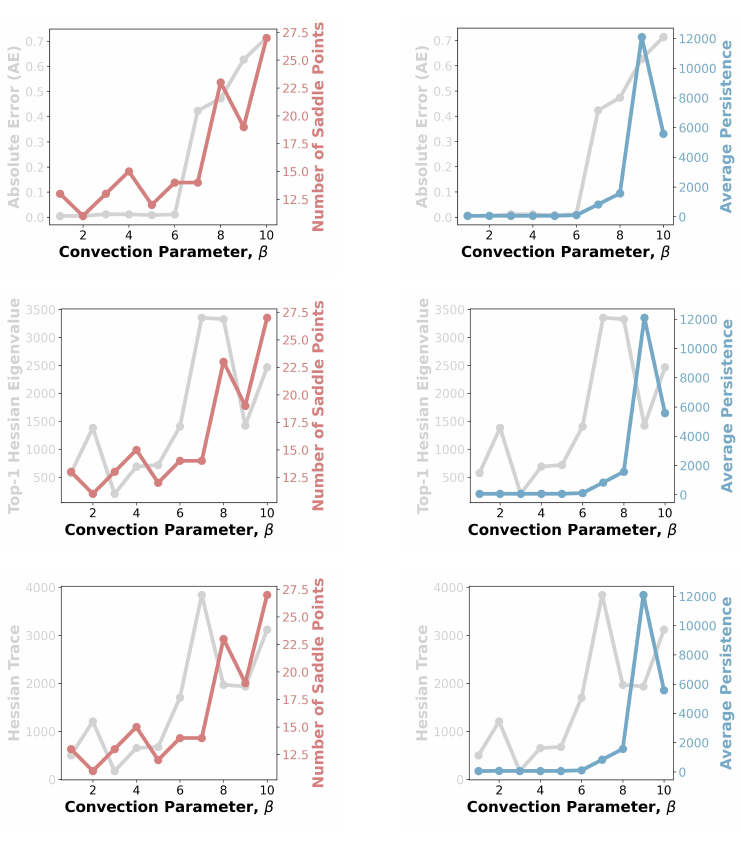}
  \caption{Quantifying the failure modes of PINNs.
    Here we show additional experimental results for PINN models solving increasingly difficult convection problems (Eq.~\ref{eq:PINN_3} for $\beta \in [1..10]$) using a fixed random seed ($seed=0$). We show our proposed TDA-based metrics, including the number of saddle points in the merge tree and average persistence, along with more traditional ML-based metrics, including the Absolute Error, Top-1 Hessian Eigenvalue, and Hessian Trace. These plots provide additional insights beyond the qualitative changes in the loss landscapes we observed in Fig.~\ref{fig:pinn}, confirming that the number of saddle points in the merge tree (left column) increases, along with the average persistence (right column), as the value of $\beta$ increases. In each row, we overlay our TDA-based metrics with a different ML-based metric. We observe similar trends in the Absolute Error, Top-1 Hessian Eigenvalue, and Hessian Trace as $\beta$ increases. Together, these results reaffirm previous findings and provide new insights into the failure modes of PINNs, suggesting that the topology of the loss landscape becomes significantly more complex and difficult to optimize.
  }
  \label{fig:experiments_pinn_scatter_plots}
\end{figure*}


\end{document}